# Learning Fault-tolerant Speech Parsing with SCREEN

Stefan Wermter and Volker Weber
University of Hamburg, Computer Science Department
Vogt-Kölln-Straße 30, D-22527 Hamburg, Germany
wermter@informatik.uni-hamburg.de
weber@informatik.uni-hamburg.de

## Abstract

This paper describes a new approach and a system SCREEN[1] for fault-tolerant speech parsing. Speech parsing describes the syntactic and semantic analysis of spontaneous spoken language. The general approach is based on incremental immediate flat analysis, learning of syntactic and semantic speech parsing, parallel integration of current hypotheses, and the consideration of various forms of speech related errors. The goal for this approach is to explore the parallel interactions between various knowledge sources for learning incremental fault-tolerant speech parsing. This approach is examined in a system SCREEN using various hybrid connectionist techniques. Hybrid connectionist techniques are examined because of their promising properties of inherent fault tolerance, learning, gradedness and parallel constraint integration. The input for SCREEN is hypotheses about recognized words of a spoken utterance potentially analyzed by a speech system, the output is hypotheses about the flat syntactic and semantic analysis of the utterance. In this paper we focus on the general approach, the overall architecture, and examples for learning flat syntactic speech parsing. Different from most other speech language architectures SCREEN emphasizes an interactive rather than an autonomous position, learning rather than encoding, flat analysis rather than in-depth analysis, and fault-tolerant processing of phonetic, syntactic and semantic knowledge.

## Introduction and Motivation

In the past, the analysis of spontaneous speech utterances as syntactic and semantic case frame representations received relatively little attention. Although there had been some early attempts for combination (Erman et al. 1980) the restricted speech and language techniques at that time forced each field, speech and language processing, to concentrate on developing further techniques separately. Therefore, in the last decade there have been primarily isolated modular attempts to build speech analyzers (e.g., (Lee, Hon, & Reddy 1990; McClelland & Elman 1986)) or language analyzers (e.g., (Hobbs et al. 1992; Kitano & Higuchi 1991)).

However, recent approaches attempt to integrate speech and language earlier to reduce the extensive space of acoustic, syntactic and semantic hypotheses (Pyka 1992; Young et al. 1989). The MINDS system (Young et al. 1989) is a speech language system which combines a speech recognizer (Lee, Hon, & Reddy 1990) with expectation-driven language analysis. The main contribution of the MINDS system is its early integration of speech hypotheses with language hypotheses in order to restrict the search space for speech processing. On the other hand, the MINDS system relies heavily on *hand-coded* pragmatic knowledge from a single domain.

The ASL system (e.g. (Pyka 1992)) is a speech language system which focused on the examination of interactions in a very general architecture. This system has an architecture similar to a blackboard architecture but without explicit control. Autonomous components can send and receive hypotheses, but the overall architecture and relationships between the components are flexible. While the MINDS system emphasized the use of pragmatic knowledge for supporting speech processing, the ASL system focused rather on syntactic and semantic knowledge. The ASL system has an extremely flexible architecture which can avoid early mistakes in favoring a particular architecture. On the other hand, this flexibility also requires very sophisticated communication operations for complexer interactions.

Both MINDS and ASL belong to the state-of-the-art architectures in speech language systems. However, in both systems the language knowledge is basically *manually encoded* and *domain-dependent*. Furthermore, currently errors like false starts, hesitations, corrections, and repetitions have only been implemented in a rudimentary pragmatic manner in the MINDS system. We designed SCREEN as a system for learning fault-tolerant incremental speech parsing. SCREEN deals with repairs (Levelt 1983), false starts, hesitations, and interjections. Since connectionist techniques have inherent fault tolerance and learning capabilities we explore these properties in a hybrid connectionist

---

[1]SCREEN stands for Symbolic Connectionist Robust EnterprisE for Natural language

architecture. In this *hybrid connectionist* architecture we make use of learning connectionist representations as far as possible, but we do not rule out symbolic representations since they may be natural and efficient for some subtasks (e.g. for testing lexical equality of two words).

The data we currently use come from the German Regensburg corpus[2] which contains dialogs at a railway counter (more than 48000 words). As a first step we used transcribed real utterances of the Regensburg corpus for SCREEN. This corpus contains a great deal of spoken constructions and occurring errors. In general we also have to deal with other errors introduced by the speech recognizer. However, for the purpose of this paper we concentrate on transcribed real speech utterances in order to illustrate the screening approach for speech parsing but our overall architecture SCREEN has the long-term goal of using speech input directly.

In this paper we will first show the underlying principles of fault-tolerant speech parsing in SCREEN and the overall architecture. Then we will describe results from flat syntactic analysis with a hybrid connectionist architecture using spoken utterances.

## Principles of fault-tolerant speech parsing with SCREEN

Our general approach is based on incremental immediate flat analysis, learning of syntactic and semantic speech parsing, and the consideration of various forms of speech related errors. The goal for this approach is to explore the parallel interactions between various knowledge sources for learning incremental speech parsing and to provide experimental contributions to the issue of architectures for speech language systems.

**Screening approach for interpretation level:** Since speech is spontaneous and erroneous, a complete interpretation at an in-depth level will often fail due to violated expectations. Therefore, we pursue a screening approach which learns an interpretation at a flat level which is more accessible for erroneous speech parsing. In particular, the screening approach structures utterances at the phrase group level.

Previous work towards this screening approach has been described as scanning understanding in SCAN (Wermter 1992). The scanning understanding primarily focused on phrase processing while our screening approach goes further by integrating and extending speech properties into a new system SCREEN for unrestricted robust spontaneous language processing.

**Learning speech parsing:** The analysis of an utterance as syntactic and semantic case frame representations is among the most important steps for language understanding. However, in addition to semantic and syntactic understanding per se, there are two central aspects: learning and speech interaction. We examine to what extent hybrid connectionist techniques can be used for learning and integrating semantic and syntactic case frame representations for speech utterances.

**Dealing with errors:** For building a speech language system we have to consider two main sources of errors: errors at the speech level and errors at the language level. Within a real speech system, errors are based on incomplete or noisy input so that many incorrect words are detected. On the other hand, even under the assumption that a speech recognizer comes up with the correct word interpretations for an utterance, there are errors at the language level like repairs, repetitions, interjections and partially incomplete phrases and sentences (e.g., telegraphic language).

## SCREEN: A system for fault-tolerant speech parsing

SCREEN has a parallel architecture with many individual modules which communicate interactively and in parallel similar to message passing systems. There is no central control; rather messages about incremental hypotheses at the current time are sent between specified modules in order to finally provide an incremental syntactic and semantic interpretation for a speech utterance. For the realization, we use hybrid connectionist techniques. That is, we integrate connectionist representations where they can be used directly and efficiently, but we do not rule out the use of other symbolic or stochastic representations. Connectionist techniques are examined because of their favorable properties of inherent fault tolerance, learning, gradedness, and parallel constraint integration. Therefore, SCREEN is not only an approach to examine fault-tolerant speech parsing but also to test the extent to which current connectionist techniques can be pushed for building a real-world complex speech language system.

### An overview

Figure 1 shows an abstract overview about the SCREEN architecture. There are basically five parts where each part consists of several modules. Each module can have a symbolic program and a connectionist network. The description of SCREEN as five parts follows its main functionalities but does not suggest a fixed hierarchical architecture. Rather, the modules in the five parts work in parallel and can exchange messages directly.

The *speech interface part* receives input from a speech recognizer as word hypotheses and provides an analysis of the syntactic and semantic plausibility of the recognized words. This analysis can be used by the speech recognizer for further speech analysis and by the subsequent language parts for filtering only important plausible speech hypotheses for further language analysis. The *category part* receives words of an utterance and provides basic syntactic, basic semantic as well as abstract syntactic and abstract semantic categories. The *correction part* receives knowledge about words and phrases as well as their categories and provides the knowledge about the occurrence of a certain error, like a repair or repetition. The *subclause part* is responsible for the detection of subclause borders in order to distinguish different subclauses. Finally, the *case frame part* is responsible for the overall interpretation of the utterance. This part receives knowledge about abstract syntactic and semantic categories of a phrase and provides the integrated interpretation.

---
[2]For clarity the illustrated examples are shown in their English translation.

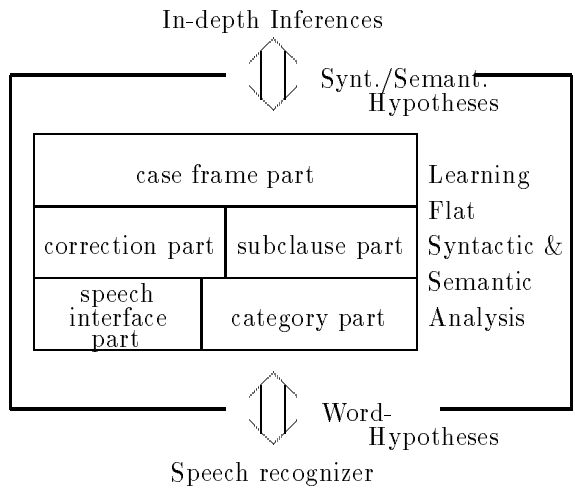

Figure 1: Overview of SCREEN

## A more detailed overview of SCREEN

Although we can not describe all the hybrid connectionist modules in SCREEN due to space restrictions, we illustrate the overall architecture and some examples for individual modules (see figure 2). We focus here only on the category part, the correction part, and the case frame part, and within these parts we will mainly focus on syntactic processing. The arrows illustrate incremental parallel flow of syntactic/semantic hypotheses. All modules in the same part in figure 2 are able to work in parallel while the processing of an utterance is incremental. While the modules in the *correction part* analyze a certain word x the modules in the *category part* are able to analyze the next word x+1 and so on.

The *category part* consists of the modules for disambiguating basic categories and determining abstract categories. The module BAS-SYN-DIS (BAS-SEM-DIS) disambiguates syntactic (semantic) basic categories. SYN-PHR-START (SEM-PHR-START) determines the start of a new syntactic (semantic) phrase group. The assignment of abstract syntactic (semantic) categories is performed by the module ABS-SYN-CAT (ABS-SEM-CAT).

The goal of the *error part* is to detect errors at a sub-word level, word level, or phrase group level. At the sub-word level the module PAUSE? checks if a current input is a pause, INTERJECTION? checks whether it is an interjection or unknown phonetic input. At the word level LEX-WORD-EQ? checks if the current word is lexically equal to the previous word and BAS-SYN-EQ? (BAS-SEM-EQ?) if it is syntactically (semantically) equal to the previous word. The modules at the phrase level are similar to the modules at the word level. LEX-START-EQ? checks if the lexical start of two phrases is equal. ABS-SYN-EQ? (ABS-SEM-EQ?) checks if the abstract syntactic (semantic) category of a current phrase group is equal to the category of the previous phrase group. The output of the modules of the correction part described so far is used in the error testing modules PAUSE-ERROR?, WORD-ERROR?, and PHRASE-ERROR?. PAUSE-ERROR? checks if a pause, interjection, or unknown phonetic input occurred, and WORD-ERROR? (PHRASE-ERROR?) determines if there is evidence for a repair at the word level (phrase group level).

In the *case frame part* a frame is filled corresponding to the syntactic and semantic categories of constituents. The module SLOT-FINDING is used to find the appropriate slot for a current phrase group. SLOT-ERROR? tests if the proposed slot is possible based on the compatibility of abstract syntactic and semantic categories for a current phrase group. VERB-ERROR? checks if new frames have to be generated. INTERPRETATION is needed to convert the internal word-by-word message structure of SCREEN to a more structured representation useful for further high level processing.

For illustration we focus on just a few modules for flat syntactic parsing. The interface of a module is represented symbolically, the learning part of a module is supported by a connectionist network. While not all modules have to contain connectionist networks they will be used as far as possible for automatic knowledge extraction. For illustrating the learned performance of some modules of SCREEN, table 1 shows three modules with a simple recurrent network SRN (Elman 1990), the number of units in the input I, hidden H, and output O layer. Training (testing) was performed with 37 (58) utterances with 394 (823) words. We used the training instances (words) based on the complete real world utterances including the errors. Under the assumption of more regular than erroneous language the general regularities will have been picked up by the network, even if it has been trained with the erroneous real-world data. For instance 99% (93%) of the basic syntactic categories of the training (test) set could be assigned correctly (see figure 2). The last row describes the combined overall performance of the modules BAS-SYN-DIS and ABS-SYN-CAT; only if both SRN-networks provide the desired category with maximum output activation it is counted as a correct combined output.

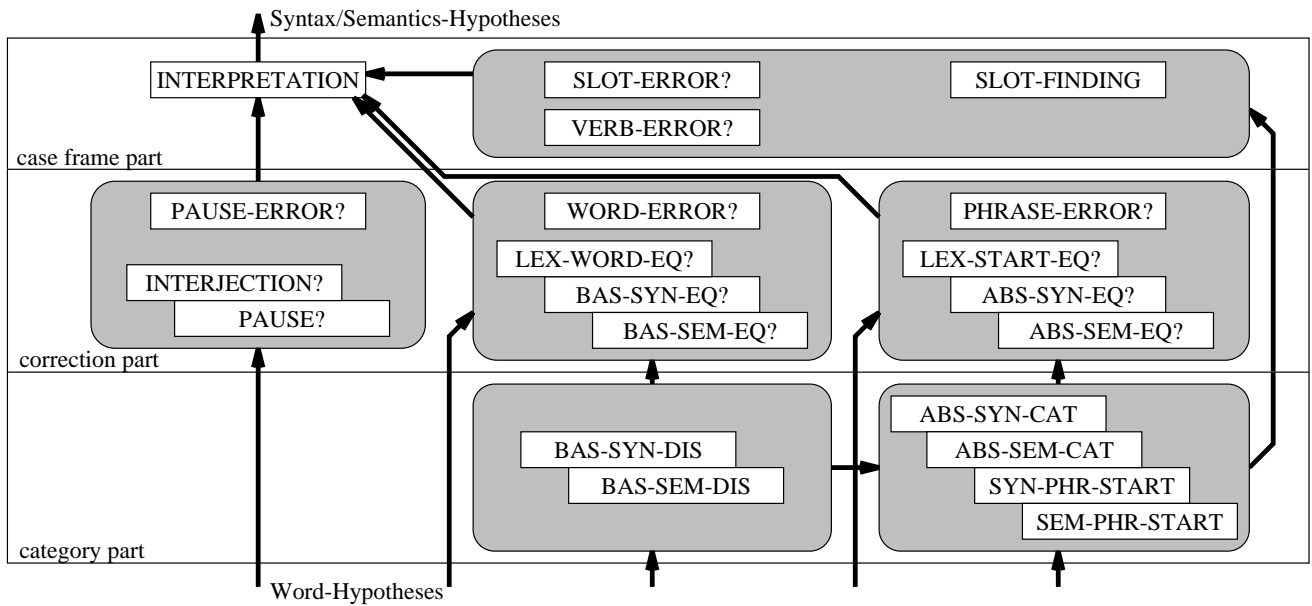

Figure 2: SCREEN: some modules of the category, correction, and case frame parts

| Module | No. of units | | | correct assignments | |
|---|---|---|---|---|---|
| | I | H | O | train | test |
| BAS-SYN-DIS | 13 | 14 | 13 | 99% | 93% |
| ABS-SYN-CAT | 13 | 7 | 8 | 91% | 85% |
| SYN-PHR-START | 13 | 7 | 1 | 93% | 89% |
| Combined | - | - | - | 90% | 82% |

Table 1: Performance of some modules

### An example for speech parsing

In this section we describe the incremental flat syntactic processing using two real transcribed utterances in SCREEN. The first sentence in figure 3 does not contain a repair, while the second in figure 4 does. The first sentence starts with the word "Yeah" which is classified as adverb by the module BAS-SYN-DIS and as part of modus group[3] by ABS-SYN-CAT. At the beginning of an utterance SYN-PHR-START classifies a word as start of a new phrase group. The second word "I" is classified as a pronoun, is part of a noun group, and starts a new phrase group. The comparison of the first word (resp. first phrase group) and second word (resp. second phrase group) does not result in any hints for a pause-, word-, or phrase error. Later in the utterance the ABS-SYN-EQ? module finds that the two syntactic phrase groups "from Regensburg" and "to Dortmund" are syntactically equal. But syntactic equality of two phrase groups alone is too weak to determine a phrase error since other modules (LEX-START-EQ? and ABS-SEM-EQ?) suggest that these two phrase groups are different with respect to their start and abstract semantic categories. When the pause "." occurs the module PAUSE-ERROR? is triggered and the pause is deleted.

For this first utterance the analysis has been rather straightforward while in the next utterance (see figure 4) we describe a more difficult example with error corrections. PAUSE-ERROR? is responsible for deleting pauses, interjections, and phonetic material. BAS-SYN-DIS classifies almost all interjections and phonetic material correctly. Only "[u]" is misclassified as adverb rather than interjection in BAS-SYN-DIS. PAUSE-ERROR? does not use this adverb information but only the output of PAUSE? and INTERJECTION?. As the module PAUSE-ERROR? determines these errors, interjections and pauses are deleted incrementally so that the phrase groups "at Monday" and "at Monday" follow each other directly. Since both groups are prepositional groups and since they have the same lexical start the modules LEX-PHRASE-EQ? and ABS-SYN-EQ? trigger PHRASE-ERROR?. Therefore the first phrase group "at Monday ..." is replaced by just "at Monday". Similarly, other types of repairs (e.g. "at Monday" replaced by "at Tuesday", "in the morning") will be dealt with in the future.

### Overall functionality and performance

SCREEN provides a fault-tolerant interpretation of a potentially faulty utterance. The words of the fault-tolerant interpretation of the faulty utterance have been underlined in order to illustrate this function-ality in figures 3 and 4. Currently corrections occur most reliably for interjections, pauses, unknown words, and syntactic repairs with lexical equality of phrase starts (as "at Monday" and "at Monday morning" in figure 4). On the other hand, an example for a currently existing undesired interpretation is "eh . in the morning at ten . in any case not after . not before nine". In this case "after" should be replaced by "before nine". However, these two prepositional phrases do not follow each other directly but there is an additional separating "not". Currently SCREEN can only deal with phrase repairs which follow each other directly since such repairs occur much more often (Levelt 1983). However, considering interjections, pauses, unknown input, and simple forms of syntactically detectable repairs in our 95 utterances we currently reach a *desired overall interpretation of 93%*.

---
[3]interrogative pronouns and confirmation words

| FAULTY/FAULT-TOLERANT UTTERANCE | BAS-SYN-DIS | ABS-SYN-CAT | SYN-PHR-START |
|---|---|---|---|
| Yeah | A | MG | |
| I | U | NG | |
| need | V | VG | |
| a | D | NG | |
| train | N | NG | |
| from | R | PG | |
| Regensburg | N | PG | |
| to | R | PG | |
| Dortmund | N | PG | |
| via | R | PG | |
| Koeln | N | PG | |
| . | - | PG | |
| with | R | PG | |
| at least | J | PG | |
| two | M | PG | |
| hours | N | PG | |
| time | N | PG | |
| in | R | PG | |
| Koeln | N | PG | |

| | |
|---|---|
| ■ □ | positive and negative activation |
| size | strength of activation |
| **A**djective, **A**dverb, **C**onjunction, **D**eterminer, **I**nterjection, Nu**m**eral, **N**oun, **P**reposition, Prono**u**n, **V**erb, - Pause | |
| **C**onjunction **G**roup, **I**nterjection **G**roup, **M**odus **G**roup, **N**oun **G**roup, **P**repositional **G**roup, **S**pecial **G**roup, **V**erb **G**roup | |

Figure 3: Syntax part of a sample parse of a sentence

| FAULTY/FAULT-TOLERANT UTTERANCE | BAS-SYN-DIS | ABS-SYN-CAT | SYN-PHR-START |
|---|---|---|---|
| when | A | MG | |
| leaves | V | VG | |
| please | V | VG | |
| . | - | IG | |
| [eh] | I | IG | |
| a | D | NG | |
| train | N | NG | |
| . | - | IG | |
| from | R | PG | |
| Regensburg | N | PG | |
| to | R | PG | |
| Dortmund | N | PG | |
| . | - | IG | |
| at | R | PG | |
| Monday | N | PG | |
| [mm] | I | CG | |
| [ts] | I | IG | |
| [u] | A | SG | |
| . | - | IG | |
| at | R | PG | |
| Monday | N | PG | |
| . | - | IG | |
| morning | A | PG | |

Figure 4: Sentence with corrections

## Discussion

We have described a screening approach to fault-tolerant speech parsing based on flat analysis. A screening approach can particularly support learning and robustness, which are properties that previous approaches did not emphasize (Young *et al.* 1989). The use of flat representations should stimulate further discussion since, in contrast to more traditional speech language systems which used highly structural hand-coded parsers, we use less structure but support fault tolerance and learning better. Therefore, speech parsers based on a screening flat analysis, learning, and fault tolerance should be more scalable, adaptive and more domain-independent.

Our approach to speech parsing is new since it makes new contributions to general architectures for speech parsing as well as new contributions to the hybrid connectionist techniques being used. With respect to the architecture we suggest a modular but interactive parallel architecture where modules exchange messages about incremental hypotheses without a particular control interpreter. With respect to the techniques we proposed the use of hybrid connectionist representations. While certain subtasks (like the symbolic equality detection of incorrectly repeated words) can be realized best using symbolic techniques, there are other subtasks with incompletely known functionality where fault-tolerant connectionist learning is advantageous.

The work which is closely related to ours is the connectionist PARSEC parser for conference registrations (Jain 1992), the hybrid connectionist JANUS speech translation system (Waibel *et al.* 1992), and the hybrid connectionist SCAN system for general phrase analysis (Wermter 1992). In general, connectionist techniques in PARSEC, JANUS, SCAN and SCREEN particularly support learning necessary knowledge where possible. However, SCREEN focuses more on exploring interactive parallel architectures and more on modeling fault tolerance.

Currently, the overall architecture as well as all the syntactic modules in SCREEN have been fully implemented, trained, and tested for a corpus of utterances with 1200 words. Although the overall SCREEN project is at an intermediate stage we believe the new architecture and the finished syntactic modules contribute substantially to new fault-tolerant learning architectures for speech language systems. Further work will focus on additional semantic modules for fault-tolerant case-role assignment and the top down interactions to speech modules in order to reduce the search space of speech hypotheses.

## Conclusion

We have described the architecture and implementation of a new speech parser which has a number of innovative properties: *the speech parser learns, it is parallel and fault-tolerant, and it directly integrates incremental processing from speech into language processing using flat analysis.* We have illustrated the processing in SCREEN with flat syntactic analysis, but in a similar way we are currently pursuing a flat semantic analysis. On the one hand, flat analysis can provide a parallel shallow processing in preparation for a more in-depth analysis for high-level dialog understanding and inferencing. On the other hand, flat analysis can potentially provide necessary restrictions for reducing the vast search space of word hypotheses of speech recognizers. Therefore, learned flat analysis in a screening approach has the potential to provide a new important intermediate link in between in-depth processing of complete dialogs and shallow processing of speech signals.

## Acknowledgements

This research was supported in part by the Federal Secretary for Research and Technology (BMFT) under contract #01IV101A0 and by the German Research Community (DFG) under contract DFG Ha 1026/6-1. We would like to thank Matthias Löchel, Manuela Meurer and Ulf Peters for their assistance with labeling the corpus and training various networks.